\documentclass{article}

\PassOptionsToPackage{square,numbers}{natbib}
\usepackage[dblblindworkshop, final]{neurips_2025}
 \usepackage{graphicx}
 \usepackage{float}
 \usepackage{placeins}
 \usepackage{bm}
 \usepackage{subcaption}
 \usepackage{comment}

\usepackage[utf8]{inputenc} 
\usepackage[T1]{fontenc}    
\usepackage{hyperref}       
\usepackage{url}            
\usepackage{booktabs}       
\usepackage{amsfonts}       
\usepackage{nicefrac}       
\usepackage{microtype}      
\usepackage{xcolor}         
\usepackage{siunitx}
\usepackage{graphicx}

\title{Grounding Foundational Vision Models with 3D Human Poses for Robust Action Recognition}
\workshoptitle{SPACE in Vision, Language, and Embodied AI}

\author{
  Nicholas Babey  \\
  Arizona State University \\
  \texttt{nbabey@asu.edu} \\
  \And
  Tiffany Gu \\
  Emory University \\
  \texttt{tiffany.gu2@emory.edu} \\
  \AND
  Yiheng Li \\
  University of California, Berkeley \\
  \texttt{leo3219@berkeley.edu} \\
  \And
  Cristian Meo \\
  LatentWorlds AI\\
  Delft University of Technology \\
  Algoverse AI Research\\
  \And
  Kevin Zhu \\
  Algoverse AI Research \\
  \texttt{kevin@algoverse.us} \\
}

\begin{document}

\maketitle

\begin{abstract}
  For embodied agents to effectively understand and interact within the world around them, they require a nuanced comprehension of human actions grounded in physical space. Current action recognition models, often relying on RGB video, learn superficial correlations between patterns and action labels, so they struggle to capture underlying physical interaction dynamics and human poses in complex scenes. We propose a model architecture that grounds action recognition in physical space by fusing two powerful, complementary representations: V-JEPA 2's contextual, predictive world dynamics and CoMotion's explicit, occlusion-tolerant human pose data. Our model is validated on both the InHARD and UCF-19-Y-OCC benchmarks for general action recognition and high-occlusion action recognition, respectively. Our model outperforms three other baselines, especially within complex, occlusive scenes. Our findings emphasize a need for action recognition to be supported by spatial understanding instead of statistical pattern recognition. Code is provided at \texttt{https://github.com/nbabey20/groundactrec}

\end{abstract}
\section{Introduction}
The ability to recognize and understand human actions is the cornerstone of embodied AI, enabling applications from collaborative robotics to assistive technologies \cite{AARNO2008692, 6696368, hu2024embodied}. However, understanding requires not only labeling an activity but also grounding the action in the occupied 3D space \cite{rajasegaran2023benefits3dposetracking, CHENG2024104991, 9022822, AGGARWAL201470}. A model must differentiate between "giving a high-five" and "reaching for an object," actions that might appear visually similar but are defined by distinct spatial and postural dynamics \cite{doering2023gated}.

Current action recognition approaches are primarily divided into two dominant methodologies: RGB-based models that utilize video pixel data to capture contextual appearance information and skeleton-based models that rely on human 3D joint data to capture movement dynamics \citep{simonyan2014two, wang2016temporal, carreira2017i3d, yan2018stgcn, shi2019two, hang2022spatial, zhang2023comprehensive}. Self-supervised video models, like V-JEPA 2 \citep{assran2025vjepa2selfsupervisedvideo}, have demonstrated prowess in understanding and predicting world states from visual data alone \cite{arnab2021vivit}. However, these RGB video models learn spatial relationships from pixel presentations, so their understanding is limited in occluded scenes where key limbs are hidden from view \cite{al2020review, yuan2023occlusions}. Conversely, skeleton-based models like CoMotion \citep{newell2025comotionconcurrentmultiperson3d}, provide explicit representations of human posture by tracking detailed 3D skeletons through visual noise and occlusion. Yet, this approach is limited by its lack of rich, contextual information like environmental cues and human-object interactions \cite{duan2022revisiting, li2024survey}. Due to the limitations of these methodologies, neither approach alone is sufficient for a spatially-grounded understanding of human action.

The limitations of these two paradigms and their complementary nature emphasizes the need for a unified model architecture that pairs contextual video understanding with precise geometric skeletons. This project is motivated by the hypothesis that fusing these complementary data streams will lead to a more robust and accurate action recognition system. This grounded understanding of how a body is configured and acting within a 3D space is critical for any embodied agent that must navigate human-centric environments.

To achieve this fusion architecture, we integrate a stream of V-JEPA 2's contextual visual features with a stream of CoMotion's 3D skeletal poses through a cross-attention mechanism. This mechanism enables each feature stream to inform one another, enforcing a holistic understanding of the action space \citep{DBLP:journals/corr/abs-2107-00135, yan2025atfusionalternatecrossattentiontransformer}. We evaluate our model’s action recognition capabilities on the Industrial Human Action Recognition Dataset (InHARD) \citep{9209531} and UCF101's \citep{soomro2012ucf101dataset101human} high occlusion subset UCF-19-Y-OCC \citep{grover2023revealing} against strong unimodal and state-of-the-art baselines. Our experiments, including a targeted ablation study on different fusion mechanisms, validate our fusion architecture's ability to ground human action within its spatial context. 

We summarize our contributions as follows: (1) A novel modality fusion that achieves physically grounded action recognition by synergizing predictive representations of a high-level world model with precise geometric data from a 3D human pose tracker, (2) A clear demonstration that fusing an implicit world model with an explicit skeletal model achieves a more spatially aware representation of human actions in highly occluded environments, and (3) A contribution to the embodied AI community by providing a model that better understands the geometric and physical nature of human interactions, a crucial step towards developing more intelligent and capable agents.

\section{Related works}

Recently, video understanding has been reshaped with the emergence of more sophisticated architectures, such as 3D convolutions \cite{carreira2017i3d, tran2018closer} and temporal transformers \cite{arnab2021vivit, bertasius2021timesformer}, that have enhanced spatio-temporal feature extraction \cite{meo2024bayesian}. A shift has been towards a large-scale self-supervised pre-training framework, enabling vast learning of general all-purpose features \citep{weinzaepfel2023crocoselfsupervisedpretraining3d, oquab2024dinov2learningrobustvisual, siméoni2025dinov3}. Models with this self-supervised framework \cite{geng2023videomae, xu2024internvideo2} use these features for action recognition \citep{qian2024advancinghumanactionrecognition}. While these models excel at general motion understanding, their reliance on pixel data can make them susceptible to challenges posed by cluttered environments where precise spatial reasoning is required \cite{meo2024object, dave2021missing, wei2024latentmaskedimagemodeling}.

Moreover, limitations of single-modality action recognition approaches have motivated the integration of different data types \cite{shin2024comprehensive, newell2025comotionconcurrentmultiperson3d, xie2025mafnet} to improve spatial understanding. These multi-modal approaches commonly fuse visual data with other input, such as depth maps \cite{shin2024comprehensive}, language models \cite{chen2024grounding, deng2025egocentric}, and skeletal data \citep{zhu2022skeletonsequencergbframe, zheng2025snndrivenmultimodalhumanaction}. By correlating visual cues with complementary data, the models rely on pattern recognition and thus do not understand the implicit physical dynamics within an action space. Our approach extends this direction by fusing the predictive capabilities of a high-level world model \cite{meo2024masked, assran2025vjepa2selfsupervisedvideo} with explicit, low-level human skeletal data \citep{newell2025comotionconcurrentmultiperson3d}. We introduce this fusion to ground actions within their contextual space so that an enriched, holistic understanding of this space is learned.

\section{Methodology}

\textbf{Feature extraction and temporal alignment}. To generate the visual feature sequence \( F_{V}\), we first sample 64 frames uniformly from each InHARD and UCF101 video clip using the Temporal Segment Network (TSN) sampling strategy \citep{wang2016temporal}. Each \(T\) frame is independently processed by V-JEPA 2's ViT-g$_{384}$ encoder that provides physical world dynamics. The transformer's [CLS] token corresponding to each frame is extracted and used as that frame's representative feature vector. This procedure results in a visual feature sequence \( F_{V} \in \mathbb{R}^{T \times D_V} \) where \(T=64\) is the number of time tokens, and \(D_{V} = 1408\) is the feature dimension.
\begin{figure}[htbp] 
  \centering
  \includegraphics[width=0.9\textwidth]{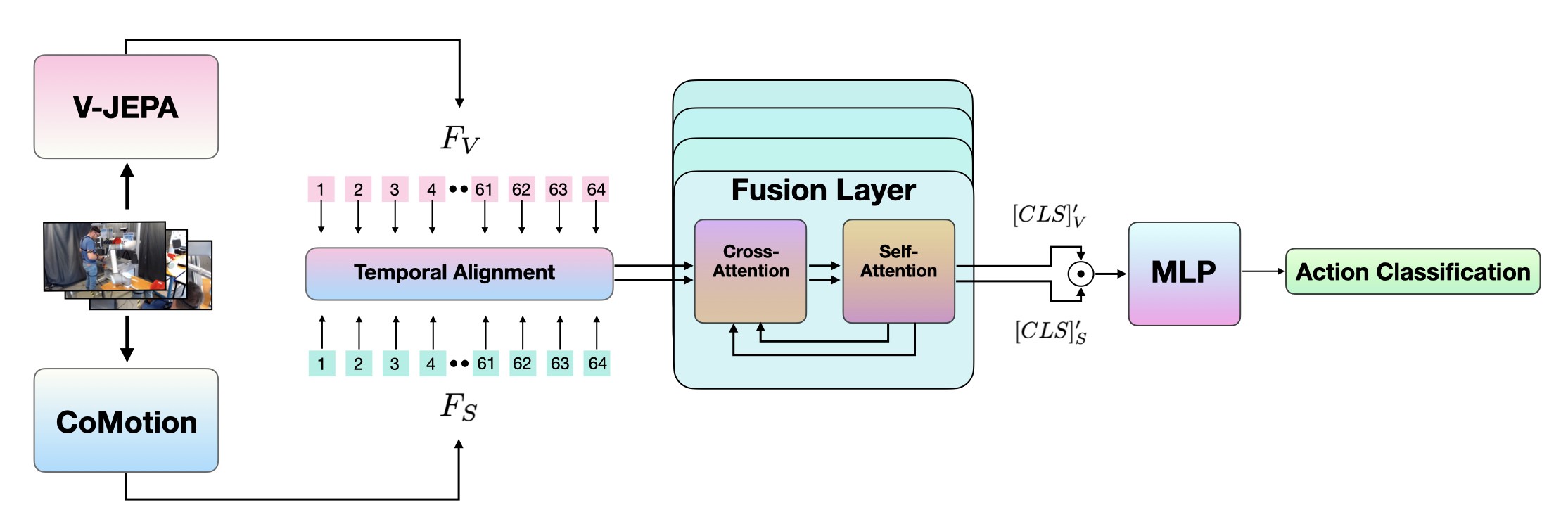}
  \caption{Fusion model architecture showing the pipeline of visual and skeletal feature sequences undergoing cross-attention and refinement to classify actions.}
  \label{fig:myplot}
\end{figure}
For the skeletal stream, CoMotion processes each frame per video clip to produce a sequence of SMPL \citep{SMPL:2015} parameters (pose \(\theta\), translation \(t\), and shape \(\beta\)) for the original clip length \(T_{clip}\). These parameters are then decoded by an SMPL layer to obtain 3D coordinates for all \(J=24\) joints that are then stored in a matrix of dimensions \([J,3]\). To ensure each representation is invariant to the person's global position, a root-relative normalization is applied by subtracting the coordinates of the root joint (pelvis at index 0) from all other joints. The normalized coordinates for each frame are converted into a feature vector by flattening the coordinate matrix into a single dimension of size \(D_S = 3J = 72\). This yields the final skeleton feature sequence \(F_S \in \mathbb{R}^{T_{clip} \times D_S}\).
To temporally align \(F_{S}\) with \(F_{V}\), we employ the same TSN sampling strategy in the visual feature extraction, enabling the fusion model to capture a holistic view of the action while enforcing a consistent input structure \citep{10.1371/journal.pone.0265115, Wang_2021_CVPR}. For a video with \(T_clip\) frames, it is divided into \(N=8\) segments of equal duration. From each segment, \(k=8\) frames are sampled to form the final sequence of \(T=n\times k= 64\). Though seemingly redundant for the visual feature stream, this process ensures the tokens from each modality correspond to the same frame to enable proper integration.
 
\textbf{Modality embedding and positional encoding}. To enable parameter sharing and preservation of modality, the temporally aligned visual features \(F_V\) and \(F_S\) are first projected into a common embedding dimension \(D_{model}=512\), using the separate linear layers: \(X_V=F_VW_V+b_V\) and \(X_S=F_SW_S+b_S\) where \(W_V \in \mathbb{R}^{D_V \times D_{model}}\) and \(W_S \in \mathbb{R}^{D_S \times D_{model}}\) are learnable weight matrices with associated bias vectors, \(b\). We prepend a learnable \([CLS]_V\) and \([CLS]_S\) token to \(X_V\) and \(X_S\), respectively, to preserve modality classification before aggregation. Finally, a sinusoidal positional encoding (\(PE \in \mathbb{R}^{(T+1) \times D{model}}\)) is added to each sequence to provide a continuous mapping of each feature sequence's temporal order. This results in updated visual and skeletal sequences represented by \(Z_{V}^{(0)}=Concat([CLS]_V, X_V) + PE\) and \(Z_{S}^{(0)}=Concat([CLS]_S, X_S) + PE\).

\textbf{Cross-attention fusion transformer}. The core architecture of our model is a stack of identical fusion layers \(L=4\) that perform bidirectional cross-attention followed by self-attention. This sequence enables mutual enrichment of each feature stream through an information exchange \citep{lin2025mvgmnstatespacemodel} followed by refinement through self-attention \citep{li2023blip2bootstrappinglanguageimagepretraining}. To achieve this, we perform a bidirectional cross-attention operation where each modality's representation is updated by attending to the other. This step consists of a multi-head attention sub-layer and an FFN, each with its own residual connection and layer normalization to stabilize training. The visual stream is updated with: \(\tilde{Z_V}=LayerNorm(Z_{V}^{(l-1)} + MultiHeadAttn(Q=Z_{V}^{(l-1)}, K=Z_{S}^{(l-1)}, V = Z_{S}^{(l-1)}\), where \(l-1\) represents output from the previous layer. An identical, parallel operation computes the updated skeleton Stream \(\tilde{Z_S}\) by swapping the roles of \(Z_V\) and \(Z_S\). These newly enriched feature representations are then passed through a standard self-attention block to enforce feature contextualization. This operation contains a multi-head attention sub-layer and an FFN to produce final layer outputs \(Z'_V\) and \(Z'_S\), where the updated visual sequence is represented by \(Z'_V=LayerNorm(\tilde{Z}_{V}+MultiHeadAttn(Q=\tilde{Z}_{V}, K=\tilde{Z}_{V}, V=\tilde{Z}_{V})\), and an identical operation is applied to update the skeletal sequence to \(Z'_S\).

\textbf{Action classification}. After the final fusion layer, the updated [CLS] tokens, \([CLS]'_{V}\) and \([CLS]'_{S}\), are extracted from their respective feature streams \(Z'_V\) and \(Z'_S\) . We concatenate these two tokens into a final feature vector that is passed through a multi-layer perception (MLP) with a softmax activation to generate action class probabilities. The full model architecture is shown in Figure \ref{fig:myplot}.

\section{Experiments}

We conduct experiments to evaluate our multimodal fusion model by exploring the following questions: (1) Does synthesizing contextual visual data with explicit skeletal data provide a more robust and spatially grounded representation of human actions? (2) Does this fusion architecture have significantly better understanding of human actions in scenes with heavy occlusion compared to the proposed baselines? To answer these questions, we validate the proposed model on action recognition using the InHARD \citep{9209531} dataset and the UCF-19-Y-OCC high occlusion dataset \citep{grover2023revealing}, and we further conducted a fusion mechanism ablation study, detailed in subsection 4.2. Our model is compared to state-of-the art V-JEPA 2, CoMotion, and fusion architecture \citep{shihata2025gatedrecursivefusionstateful} baselines. Further details on the datasets and experiment configurations are contained in \ref{appendix:dataset} and \ref{appendix:config}, respectively.

\begin{table*}[t]
\centering

\begin{minipage}{\textwidth}
\centering
\caption{Evaluation of action recognition performance on the InHARD and UCF-19-Y-OCC benchmarks. Best results per benchmark are indicated in \textbf{bold}.}
\label{tab:main-results}
\resizebox{\textwidth}{!}{
\begin{tabular}{l ccc c ccc}
\toprule
& \multicolumn{3}{c}{\textbf{InHARD}} & & \multicolumn{3}{c}{\textbf{UCF-19-Y-OCC}} \\
\cmidrule{2-4} \cmidrule{6-8}
\multicolumn{1}{c}{Model} &
\multicolumn{1}{c}{Top-1 Acc. (\%)$\uparrow$} &
\multicolumn{1}{c}{Macro mAP (\%)$\uparrow$} &
\multicolumn{1}{c}{Macro F1 (\%)$\uparrow$} & &
\multicolumn{1}{c}{Top-1 Acc. (\%)$\uparrow$} &
\multicolumn{1}{c}{Macro mAP (\%)$\uparrow$} &
\multicolumn{1}{c}{Macro F1 (\%)$\uparrow$} \\
\midrule
V-JEPA 2 baseline & $80.76 \pm 0.21$ & $80.93 \pm 0.43$ & $76.24 \pm 0.30$ && $31.83 \pm 0.76$ & $\bm{58.48 \pm 0.47}$ & $14.23 \pm 0.88$ \\
CoMotion baseline & $75.92 \pm 0.17$ & $74.60 \pm 0.41$ & $69.52 \pm 0.11$ && $6.20 \pm 0.00$ & $8.84 \pm 0.25$ & $1.72 \pm 0.16$ \\
\textbf{Fusion model (cross-attention)} & $\bm{83.47 \pm 0.03}$ & $\bm{84.96 \pm 0.10}$ & $\bm{80.21 \pm 0.31}$ && $\bm{38.62 \pm 0.22}$ & $54.10 \pm 0.14$ & $\bm{16.30 \pm 0.41}$ \\
Gated recursive fusion & $79.25 \pm 1.54$ & $76.90 \pm 0.90$ & $73.69 \pm 1.58$ && $29.54 \pm 1.78$ & $50.07 \pm 0.79$ & $11.44 \pm 0.64$ \\
\bottomrule
\end{tabular}
}

\end{minipage}

\vspace{1.5em} 

\begin{minipage}{\textwidth}
\centering
\caption{Ablation study on different fusion methods evaluated on InHARD and UCF-19-Y-OCC.}
\label{tab:main-results}
\resizebox{\textwidth}{!}{%
\begin{tabular}{l ccc c ccc}
\toprule
& \multicolumn{3}{c}{\textbf{InHARD}} & & \multicolumn{3}{c}{\textbf{UCF-19-Y-OCC}} \\
\cmidrule{2-4} \cmidrule{6-8}
\multicolumn{1}{c}{Fusion method} &
\multicolumn{1}{c}{Top-1 Acc. (\%)$\uparrow$} &
\multicolumn{1}{c}{Macro mAP (\%)$\uparrow$} &
\multicolumn{1}{c}{Macro F1 (\%)$\uparrow$} & &
\multicolumn{1}{c}{Top-1 Acc. (\%)$\uparrow$} &
\multicolumn{1}{c}{Macro mAP (\%)$\uparrow$} &
\multicolumn{1}{c}{Macro F1 (\%)$\uparrow$} \\
\midrule
Early fusion (concatenation) & $79.52 \pm 0.83$ & $78.55 \pm 1.16$ & $75.91 \pm 1.46$ && $33.34 \pm 1.10$ & $\bm{56.50 \pm 2.51}$ & $14.87 \pm 0.41$ \\
Late fusion (score averaging) & $80.24 \pm 0.53$ & $83.59 \pm 0.73$ & $77.31 \pm 0.55$ && $34.42 \pm 0.78$ & $53.2 \pm 0.25$ & $14.93 \pm 0.44$ \\
\textbf{Fusion model (cross-attention)} & $\bm{83.47 \pm 0.03}$ & $\bm{84.96 \pm 0.10}$ & $\bm{80.21 \pm 0.31}$ && $\bm{38.62 \pm 0.22}$ & $54.10 \pm 0.14$ & $\bm{16.30 \pm 0.41}$ \\
\bottomrule
\end{tabular}
}
\end{minipage}

\end{table*}

\subsection{Results}

Table 1 shows the results of each model's action recognition performance on each benchmark. We find that our fusion model outperforms the other baseline models in all three metrics, and the V-JEPA 2 baseline only has slightly lower performance compared to our model. These results emphasize the robust and spatially-grounded action recognition capabilities achieved through our modality fusion, yet they also showcase V-JEPA 2's powerful prediction and understanding capabilities. 

The right side of Table 1 shows each model's performance on the high occlusion benchmark UCF-19-Y-OCC. Our model significantly outperforms the baseline models, with a 6.79\% higher accuracy score than V-JEPA 2's baseline. Notably, CoMotion alone collapses under heavy occlusion. This performance gap shows that the fusion of a contextual vision model with explicit pose data significantly enhances human action understanding in complex scenes. To evaluate the effectiveness of our cross-attention mechanism, an ablation study to compare different fusion techniques is seen in Table 2. Our model is seen to have a slight performance advantage compared to these techniques, reflecting cross-attention's effectiveness in enriching and contextualizing complementary feature streams. 

\section{Conclusion}

In this paper, we propose a multimodal fusion architecture that effectively integrates rich, contextual representations from the V-JEPA 2 video model with precise, geometric skeleton trajectories from CoMotion. Our experiments confirm that this multimodal approach leads to a more robust and spatially grounded understanding of human action. Our model achieved superior performance on both action recognition and high-occlusion action recognition benchmarks. These findings attest to the value of complementing finer data representations within broader contextual data to enhance spatial understanding. However, limitations of our work come from its dependency on feature extraction from V-JEPA 2 and CoMotion, along with its limited testing on action recognition benchmarks and sparse availability of state-of-the-art fusion baseline performances on these benchmarks. While this research has the potential to positively impact society by advancing applications in collaborative robotics and assistive technologies, it's important to consider potential negative implications, such as malicious use of human activity monitoring in surveillance applications. Ultimately, our results advocate for a shift from statistical pattern action recognition toward recognition gained from a holistic understanding of both human interaction's and their spatial context.

\newpage
\bibliographystyle{ieeetr}
\bibliography{references}

\clearpage
\appendix
\section{Appendix}

\subsection{Dataset details}
\label{appendix:dataset}
Our experiments are conducted on the Industrial Human Action Recognition Dataset (InHARD) that contains over two million frames collected from 16 different subjects over 13 different industrial action classes. This benchmark is designed from complex, interaction-heavy human activities that feature frequent occlusions and presents a challenge for models that do not effectively reason about spatial and postural information. The dataset provides 14 meta action labels, which we use as the standard for training and evaluation. 

A critical pre-processing step in our methodology is the cropping of the InHARD video clips. The raw videos in the dataset are presented as mosaics of three synchronized camera views:top-left, top-right, and bottom-right, with an empty bottom-left quadrant. Therefore, we cropped each of these views into separate video clips that created singular left, right, and top views for each InHARD video. The cropping ensures actors remain centered so that CoMotion, which heavily relies on visual cues from body joints, can more accurately detect SMPL joints. The cropping presents clear, per-frame skeletons of the actor, so occlusion patterns can be properly detected to eventually create an occlusion-heavy subset of InHARD for model evaluation.

To specifically test our model's performance in complex scenes with occluded interactions, we use the UCF-19-Y-OCC \citep{grover2023revealing} occlusion benchmark, which is a subset of the UCF-101 \citep{soomro2012ucf101dataset101human} action recognition benchmark. This benchmark was manually curated for action recognition performance under real-world occlusion. It consists of 1,732 video clips that cover 19 action classes that frequently involve natural occlusions (environmental clutter, challenging camera angles, and limbs blocked by objects). Unlike other synthetic occlusion datasets, UCF-19-Y-OCC provides a realistic benchmark for evaluating a model's ability to generalize and maintain performance when presented with incomplete visual evidence in a complex environment. 

\subsection{Experiment configuration}
\label{appendix:config}
For our experiments involving InHARD, we train all models on the cropped dataset and use its official train and validation splits across the 14 meta action labels. For our high occlusion experiment, each model is trained on the first UCF101 training split, and they are evaluated on the UCF-19-Y-OCC high occlusion benchmark. Each model is trained on a RunPod NVIDIA A100 SXM GPU for 30 epochs, employing the AdamW optimizer with a learning rate of \(3 × 10^{-4}\) and a weight decay of 0.05 for the attention probes. The learning rate was managed by a cosine decay schedule with a 5\% warmup period. The training set was split into batches of size 128. To handle variability in \(T_{clip}\), all models employ TSN with the previously described protocol of being divided into eight clip segments each containing eight feature vectors. 

To ensure stable training, we utilize a dropout rate of 0.1, gradient clipping at a max norm of 1.0, and AMP to accelerate computation. The baseline attentive probes are configured with a model dimension \(D_{model}\) of 1408 for the V-JEPA 2 baseline and 256 for the CoMotion baseline while both have two transformer layers and eight attention heads. Our fusion model has a \(D_{model} = 512\), with four fusion layers and eight attention heads. The best-performing checkpoint for each model is selected based on the highest mean Average Precision (mAP) on the validation set for each run. Three different seeds are tested to ensure experiment variability, and the mean values with standard deviation are presented in the tables. 

\subsection{Ablation study details}
\label{appendix:ablation}
To evaluate our model's cross-attention fusion mechanism, we conduct an ablation study against two other fusion models: early fusion via feature concatenation, and late fusion via averaging prediction scores. For a fair comparison, all other model components and training hyperparameters were held constant across experiments. The performance of each fusion strategy was evaluated on the InHARD benchmark using the same evaluation metrics as the first experiment. The results clearly demonstrate that Cross-Attention fusion consistently outperforms both early and late fusion, highlighting its superior capability at modeling complex relationships between fused features and action recognition. 

\section{Extended Related Works}
\label{appendix:related}
A complementary line of research focuses on explicit representations of human movement. Recent progress in 3D human pose estimation has enabled the recovery of detailed skeletal and body mesh information from a single camera \cite{zhang2022pymaf}. These methods can track multiple individuals through time, even when they are partially obscured, by leveraging temporal dependencies and learned priors \cite{newell2025comotionconcurrentmultiperson3d, doering2023gated}. Such skeletal data provides a robust, geometry-based understanding of posture and limb configurations, which is often difficult to infer implicitly from raw video pixels alone. Researchers have also explored action recognition solely using these skeletal streams, often employing graph convolutional networks to model the spatial and temporal relationships of the joints \cite{yan2018stgcn, li2022stfocus}.

\end{document}